\def\year{2022}\relax
\documentclass[letterpaper]{article} 
\usepackage{aaai22}  
\usepackage{times}  
\usepackage{helvet}  
\usepackage{courier}  
\usepackage[hyphens]{url}  
\usepackage{graphicx} 
\usepackage{natbib}  
\usepackage{caption} 
\DeclareCaptionStyle{ruled}{labelfont=normalfont,labelsep=colon,strut=off} 
\usepackage{algorithm}
\usepackage{algorithmic}
\usepackage{graphicx}
\usepackage{amsmath}
\usepackage{amssymb}
\usepackage{booktabs}
\usepackage{amsfonts}

\usepackage{multirow}
\usepackage{array}
\usepackage{color}
\usepackage{subcaption}
\usepackage{colortbl}
\usepackage{arydshln}
\graphicspath{{images/}}
\usepackage{bbding}
\usepackage{makecell}
\usepackage{hhline}
\usepackage{bm}

\usepackage[capitalize]{cleveref}
\crefname{section}{Sec.}{Secs.}
\Crefname{section}{Section}{Sections}
\Crefname{table}{Table}{Tables}
\crefname{table}{Tab.}{Tabs.}

\usepackage{newfloat}
\usepackage{listings}
\floatstyle{ruled}
\newfloat{listing}{tb}{lst}{}
\floatname{listing}{Listing}
\title{Self-Annotated Training for Controllable Image Captioning}
\author{
    Zhangzi Zhu, Tianlei Wang, and Hong Qu\\
University of Electronic Science and Technology of China\\
{\tt\small 202021080414@std.uestc.edu.cn}
}
\usepackage{bibentry}

\begin{document}

\maketitle

\begin{abstract}
The Controllable Image Captioning (CIC) task aims to generate captions conditioned on designated control signals. Several structure-related control signals are proposed to control the semantic structure of sentences, such as sentence length and Part-of-Speech tag sequences. However, due to the fact that the accuracy-based reward focuses mainly on contents rather than semantic structures, existing reinforcement training methods are not applicable to structure-related CIC models. The lack of reinforcement training leads  to exposure bias and the inconsistency between the optimizing function and evaluation metrics. In this paper, we propose a novel reinforcement training method for structure-related control signals: Self-Annotated Training (SAT), to improve both the accuracy and controllability of CIC models. In SAT, a recursive annotation mechanism (RAM) is designed to force the input control signal to match the actual output sentence. Moreover, we propose an extra alignment reward to finetune the CIC model trained after SAT method, which further enhances the controllability of models. On the MSCOCO benchmark, we conduct extensive experiments on different structure-related control signals and on different baseline models, the results of which demonstrate the effectiveness and generalizability of our methods.
\end{abstract}

\begin{table}[ht]
	\begin{center}
		\begin{tabular}
		{|c|p{120 pt}<{\centering}|c|c|}
		    \hline
		    \multicolumn{4}{|l|}{\hspace{-11pt} \raisebox{-.5\height}{ \includegraphics[scale=0.20]{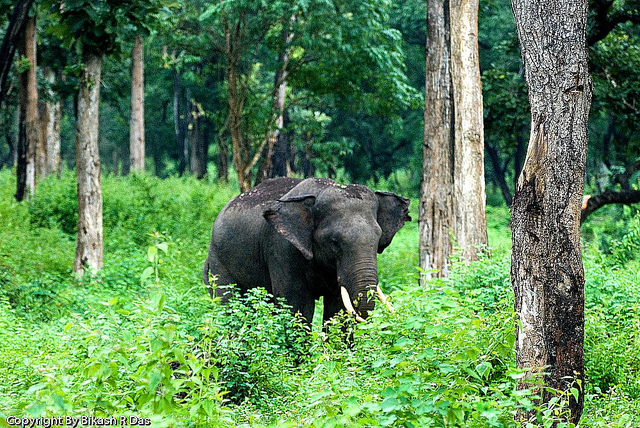} } \hspace{-5pt} \multirow{1}{*}[14 pt]{GT: an elephant walking} \hspace{-85pt} \multirow{1}{*}[4 pt]{through the weeds} \hspace{-78pt} \multirow{1}{*}[-6pt]{in the forest.} \hspace{-75pt} \multirow{1}{*}[-25 pt]{control signal: length = 9}} \\
			\hline
			method & captions & acc. & con.\\
			\hline
			\multirow{1}{*}[-8 pt]{XE} & an elephant is walking through a field of grass. & \multirow{1}{*}[-8 pt]{$\times$} & \multirow{1}{*}[-8 pt]{\checkmark} \\
			\hline
			\multirow{1}{*}[-8 pt]{RL} & an elephant walking through a forest with trees in it. & \multirow{1}{*}[-8 pt]{\checkmark} & \multirow{1}{*}[-8 pt]{$\times$} \\
			\hline
			\multirow{1}{*}[-8 pt]{SAT} & an elephant walking through the grass in the forest. & \multirow{1}{*}[-8 pt]{\checkmark} & \multirow{1}{*}[-8 pt]{\checkmark} \\
			\hline
		\end{tabular}
	\end{center}
	\caption{In the task of sentence length, we show the comparisons of structure-related CIC models trained with (a) cross-entropy training (XE), (b) conventional reinforcement training (RL), and (c) our self-annotated training (SAT). As we can see, only the model trained with our SAT method achieves both high accuracy and controllability.}
\label{Table1}
\end{table}

\section{Introduction}
Image captioning, which belongs to the intersection of computer vision and natural language processing, is an important part of applying artificial intelligence to many life scenes. The generated texts can be used for the image search task and visually impaired people assistance. Captioning generation is a particularly challenging task which requires models not only to recognize salient objects, attributes and relationships in an image, but also to describe various information through fluent natural language. Thanks to the proposal of encoder-decoder framework \cite{Vinyals2015Show} and attention mechanism \cite{2015Show}, current captioning models \cite{2020X, luo2021duallevel, 2020Improving, 2020Image} have already outperformed humans in several accuracy-based evaluation metrics.

In recent years, many efforts \cite{2021Human, Cornia2020Show, Chen2020Say, 2020Length} have been made to endow captioning models with human-like controllability, called Controllable Image Captioning (CIC). They introduce various control signals into CIC. As classified in \cite{2021Human}, control signals can be roughly divided into two categories: 1) \emph{Content-related}: the control signal is related to the contents of images, such as guiding objects \cite{Zheng2019Intention}, image regions \cite{Cornia2020Show}, abstract scene graphs \cite{Chen2020Say} and verb-specific semantic roles \cite{2021Human}. 2) \emph{Structure-related}: the control signal is related to the semantic structures of sentences, including sentence length \cite{2020Length} and Part-of-Speech (POS) tags \cite{Deshpande2019Fast}. Trained with cross-entropy loss, existing CIC models have achieved satisfactory performance in terms of controllability.

Nevertheless, current structure-related CIC works are unable to combine with reinforcement training methods \cite{Rennie2016Self, Gao2019Self}, which prevents models from generating more accurate sentences. It has been proved that models trained only with cross-entropy loss suffer from exposure bias \cite{2015Sequence}, since in the training stage, the word at each time step is generated conditioned on ground truth words while in the testing stage, the word is generated based on previously predicted words of the model. Besides, the inconsistency between the cross entropy loss in the training stage and non-differentiable evaluation metrics in the testing stage also leads to unsatisfactory results. Therefore, reinforcement training, which can solve the above problems, is crucial for structure-related CIC models.

For content-related CIC tasks, only the models in \cite{Cornia2020Show, 2021Human} are trained with REINFORCE algorithm, where control signals aligned with ground truth captions are used as inputs. Under the guidance of these content-related control signals, the content of the generated sentence is trained to approach the ground truth caption, so that CIC models are able to learn content-related controllability. However, this method can only be applied to content-related control signals since the reward in reinforcement learning is designed to measure the content similarity between two sentences. For structure-related control signals, the reward makes the generated sentence approach the ground truth sentence in contents rather than semantic structures. Therefore, CIC models fail to learn structure-related controllability with ground truth annotations during conventional reinforcement training (Fig. \ref{Table1}).

In this paper, we propose a novel reinforcement training method, \emph{Self-Annotated Training} (SAT), for structure-related control signals. The main difference between our SAT and the method in \cite{Cornia2020Show} is the source of input control signals. During reinforcement training, control signals in \cite{Cornia2020Show} come from ground truth captions, while ours are from generated sentences of CIC models. For this purpose, we design a recursive annotation mechanism (RAM) which forces the input control signal to match the actual output sentence. Moreover, we propose an extra alignment reward to finetune the CIC model trained after SAT method. Under the extra supervision of the alignment reward, the controllability of the CIC model is further enhanced. Experiments on MSCOCO dataset \cite{Lin2014Microsoft} demonstrate that SAT can effectively improve the accuracy and controllability of captioning models controlled by structure-related signals. Besides, we conduct quantitative experiments on different structure-related control signals and on different baseline models, which show the generalizability of our SAT method.

In summary, we mainly make the following contributions in this paper:

\begin{itemize}
\item We propose a novel reinforcement training method for structure-related CIC: \emph{Self-Annotated Training} (SAT), which can be easily incorporated into existing captioning models to make them generate more accurate and controllable sentences. To the best of our knowledge, SAT is the first reinforcement training method for structure-related CIC models.
\item We propose an extra alignment reward to finetune the CIC model trained after SAT method. Under the extra supervision of the alignment reward, the controllability of the CIC model is further improved.
\item We perform extensive experiments on different structure-related control signals and on different baseline models, which demonstrate the effectiveness and generalizability of our methods.
\end{itemize}

\section{Related Work}
\subsection{Image Captioning}
Existing captioning models follow an encoder-decoder framework which is first introduced to image captioning tasks by \cite{Vinyals2015Show}. In \cite{2015Show, Huang2019Attention}, attention mechanism is used to select the target area of interest that needs special attention at each time step. To enhance the diversity, GAN-based methods \cite{Dognin2019Adversarial, Dai2017Towards, Chen2018Improving} are introduced in image captioning. Models proposed in \cite{Zheng2019Intention, Ge2019Exploring} change the order of the sentence generation, starting from the middle or the end of sentences. Two-step networks are designed in \cite{2020Image, 2019Show} to generate refined captions from raw information. In \cite{Yang2018Auto, Chen2020Say, shi-etal-2020-improving}, scene graphs are employed to further explore the objects, attributes and relationships in the image, which improve the overall performance of captioning models. In order to solve the long-term dependency problem in the previous LSTM architectures, Transformer-based models \cite{luo2021duallevel, 2020Meshed, DBLP:journals/corr/abs-2003-08897, 2019Entangled} using multi-head self- and encoder-decoder attention mechanisms are explored. Regarding the training strategy, reinforcement learning methods \cite{Rennie2016Self, Gao2019Self} are proposed to optimize non-differentiable metrics, solving the problem of exposure bias. In this paper, we expand the scope of application of reinforcement training from conventional captioning tasks to CIC tasks. With our self-annotated training, the accuracy and controllability of CIC models are further improved.

\subsection{Controllable Image Captioning}
In addition to the conventional captioning task, another related route is to generate controllable captions, which is called controllable image captioning (CIC). CIC models aim to generate captions conditioned on designed control signals. As classified in \cite{2021Human}, control signals can be roughly divided into two categories: 1) \emph{Content-related}: the control signal is related to the contents of images. Models in \cite{Zheng2019Intention} generate sentences starting from a guiding object in order to contain the given word. Models in \cite{Cornia2020Show} describe images conditioned on a given sequence or set of image regions to control which objects are described and their orders. In \cite{Chen2020Say}, Abstract Scene Graphs (ASG) are taken as the control signal to control sentences at a more fine-grained level. \cite{2021Human} proposes a new control signal, Verb-specific Semantic Roles (VSR), which meets both event-compatible and sample-suitable requirements. 2) \emph{Structure-related}: the control signal is related to the semantic structures of sentences. To explore length-aware image captioning models, sentence length is studied in \cite{2020Length, 2020Controlling} as the signal of “length level”. Models in \cite{Deshpande2019Fast} employ signals of Part-of-Speech (POS) tag sequences to make generated sentences diverse. However, existing reinforcement training methods are not applicable to structure-related CIC models above. In this work, we propose the SAT method, the first reinforcement training method for structure-related signals, to achieve high accuracy and controllability of models.

\section{Preliminaries}
\subsection{Embedding Method}

\begin{table}[htbp]
	\begin{center}
		\begin{tabular}{|c|c||c|c|}
			\hhline{--||--}
			$ \beta_{len} $ & length & $ \beta_{ten} $ & tense\\
			\hhline{--||--}
			0 & $ \leq $ 8	& 5 & no v \\ \hhline{--||--}
            1 & $ = $ 9	& 6 & be + v \\ \hhline{--||--}
            2 & $ = $ 10 & 7 & v-ing \\	 \hhline{--||--}
            3 & $ = $ 11 & 8 & v \\ \hhline{--||--}
			4 & $ \geq $ 12	& 9 & v-ed \\ \hhline{--||--}
		\end{tabular}
	\end{center}
	\caption{Specific settings of control levels for sentence length and tense.}
\label{Table2}\end{table}

In this work, we conduct experiments on structure-related control signals of sentence attributes, including sentence length and tense. For each caption, we divide it into a length level $ \beta_{len} $ and a tense level $ \beta_{ten} $ according to its attribute. The specific settings of control levels for sentence length and tense are shown in Table \ref{Table2}. We try to make the number of samples at each level evenly distributed.

Take the task of controlling the sentence length and tense simultaneously as an example. Given an input caption $ Y = \{y_1, y_2, \cdots, y_T\} $, we first obtain the control signal $ \beta = \{\beta_{len}, \beta_{ten}\} $ according to the attribute of $ Y $. For each element in $ \beta $, an embedding matrix $ W\in \mathbb{R}^{k\times d} $ (k is the number of levels in Table \ref{Table2} and d is the embedding dimension) is employed to embed it to a $ d $-dimensional vector space:
\begin{equation}\label{equ1}
\begin{aligned}
e_{len} = W^T\Pi_{len}, \\
e_{ten} = W^T\Pi_{ten},
\end{aligned}
\end{equation}
where $ \Pi_{len} $ and $ \Pi_{ten} $ are the one-hot representations of $ \beta_{len} $ and $ \beta_{ten} $ respectively. After obtaining the length level embedding $ e_{len} $ and the tense level embedding $ e_{ten}  $, we calculate the control level embedding $ e_\beta $ as (remove another part when only one element is desired to be controlled):
\begin{equation}\label{equ2}
e_\beta = e_{len} + e_{ten}.
\end{equation}
Then, each word $ y_i $ in caption $ Y $ is represented by adding the control level embedding $ e_\beta $ with word embedding $ e_{y_i} $ and, optionally (for Transformer-based \cite{2017Attention} decoder), positional embedding $ e_{p_i} $:
\begin{equation}\label{equ3}
x_i = e_\beta + e_{y_i} + e_{p_i}.
\end{equation}
Finally, $ x_i $ replaces original word embedding as the input of the decoder of captioning models. By integrating the control signal information into the word embedding, CIC models naturally associate the input control signal with the output sentence, thus learning the meaning of the control signal.

\subsection{Training Strategy}
\subsubsection{Cross Entropy Training (XE)} Given an image $ I $, a target ground truth sequence $ y_{1:T}^* $, the paired control signal $ \beta $ and the captioning model with parameters $ \theta $, we minimize the following cross-entropy loss:
\begin{equation}\label{equ4}
L_{XE}(\theta) = -\sum_{t=1}^{T} log(p_\theta(y_t^*|y_{1:t-1}^*, I, \beta)),
\end{equation}
\noindent where $ \beta $ depends on the attribute of the ground truth sequence.
\subsubsection{CIDEr Score Optimization (RL)} After pretrained with cross-entropy loss, CIC models are further trained by REINFORCE algorithm. The training process is to minimize the negative expected reward:
\begin{equation}\label{equ5}
L_{RL}(\theta) = -\mathbb{E}_{y_{1:T} \sim p_\theta} [r(y_{1:T})],
\end{equation}
\noindent where the reward $ r(\cdot) $ is calculated according to the score of the evaluation metric (e.g.CIDEr \cite{Vedantam2015CIDEr}). As in \cite{Rennie2016Self}, the gradient can be approximated as:
\begin{equation}\label{equ6}
\nabla_\theta L_{RL}(\theta) \approx -\frac{1}{k}\sum_{i=1}^{k} (r(Y_i^s) - b) \nabla_\theta log(p_\theta(Y_i^s|I, \beta_i)),
\end{equation}
\noindent where $ Y_i^s $ represents the $ i $-th sampled caption and $ b = (\sum_{i} r(Y_i^s)) / k $ is the baseline, computed as the mean of the rewards obtained by the sampled captions. In this stage, the input control signal $ \beta_i $ is calculated from the attribute of the $ i $-th ground truth caption aligned with image $ I $.

\begin{figure*}[htbp]
\centering
\includegraphics[width=2\columnwidth]{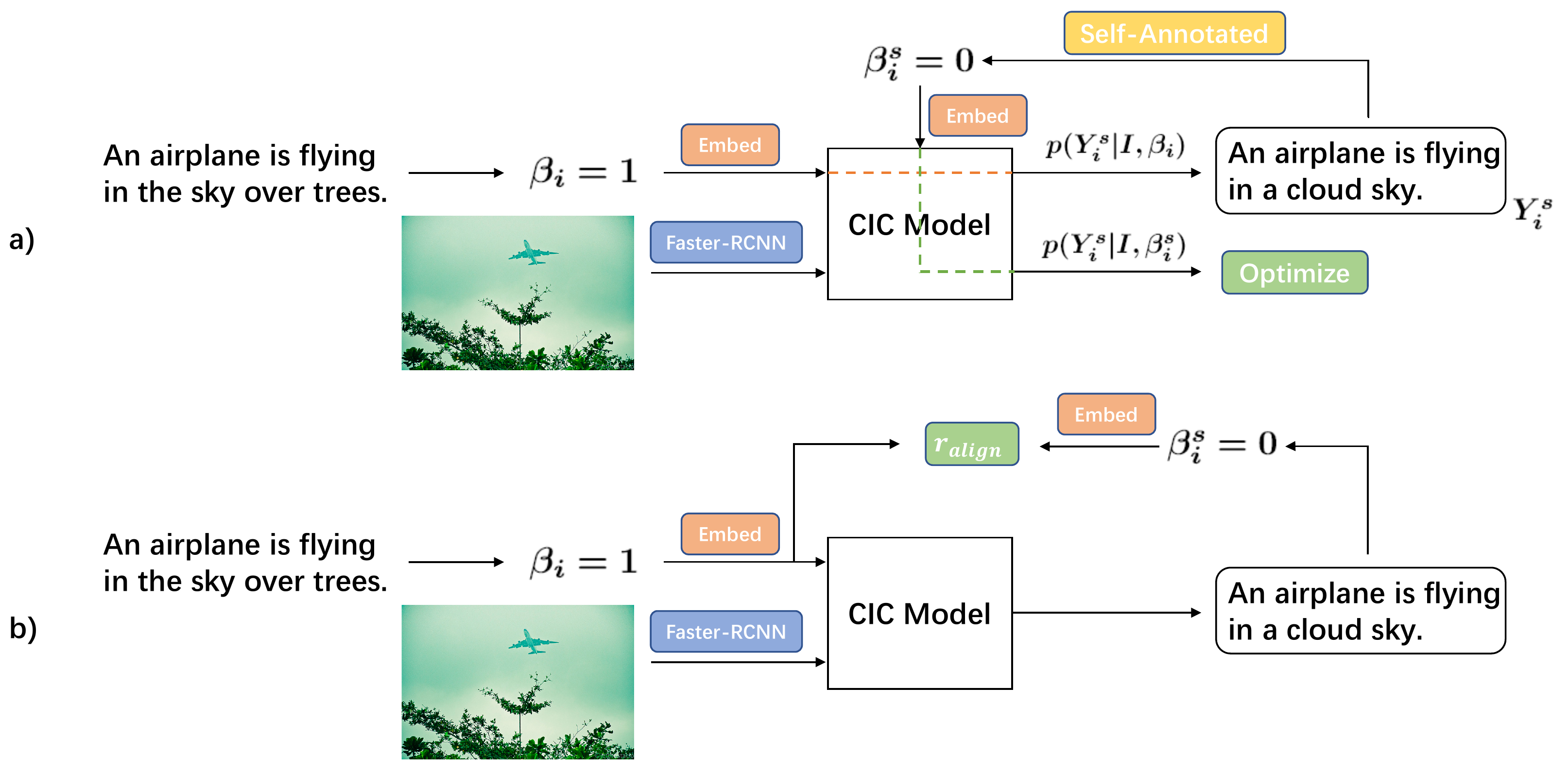} 
\caption{Overview of our (a) self-annotated training, (b) finetuning method, in the task of sentence length.}
\label{fig1}
\end{figure*}

\section{Method}
In this section, we first elaborate on the specific process of Self-Annotated Training (SAT) in \Cref{sec:formatting41}. Then, we introduce how to finetune the CIC model after SAT in \Cref{sec:formatting42}.
\subsection{Self-Annotated Training (SAT)}
\label{sec:formatting41}

By reviewing the cross-entropy training method, we can find that the main reason why models learn controllability well is the consistency of inputs and outputs, that is, the input control signal can match the output sentence of the model (during cross-entropy training output sentences of models are ground truth captions). In this case, CIC models are able to naturally associate the input control signal with the output sentence, thus learning the meaning of the control signal. Therefore, during reinforcement training, control signals should be aligned with actual output sentences which are generated by sample methods rather than ground truth captions. However, output sentences generated by sample methods require control signals as the input. The control signals, in turn, are calculated based on output sentences. They are prerequisites to each other, which brings great obstacles to the actual operation. Therefore, how to design an algorithm to solve the above problem is crucial.

Figure \ref{fig1} (a) gives an overview of our proposed SAT method. Its central idea is that the control signal is from actual sampled sentences rather than ground truth captions. Given an image $ I $ and its $ k $ paired ground truth captions $ G = \{G_1, G_2, \cdots, G_k\} $, for each ground truth $ G_i $, we first calculate the control signal $ \beta_i $ for caption $ G_i $ according to its attribute. Then, a single Monte-Carlo sample is used to generate a sampled sentence $ Y_i^s $ and its corresponding distribution $ p(Y_i^s|I,\beta_i) $ conditioned on both the input image $ I $ and the control signal $ \beta_i $. It is worth noting that the attribute of $ Y_i^s $ is not necessarily consistent with the control signal $ \beta_i $ due to the limited capability of the CIC model. Therefore, using Eq. (\ref{equ6}) directly for reinforcement training will fail to improve the controllability of the model since the fact that accuracy-based reward only evaluates the content similarity instead of structure-related consistency.

In order to force the input control signal to match the output sampled sentence, a recursive annotation mechanism (RAM) is designed. After obtaining the sampled sentence $ Y_i^s $, we compute the control signal $ \beta_i^s $ for $ Y_i^s $. Then, the model regards $ Y_i^s $ as the target caption and predicts output distributions $ p(Y_i^s|I,\beta_i^s) $ conditioned on $ I $ and $ \beta_i^s $, the process of which is the same as that in cross-entropy training. In this case, the input control signal $ \beta_i^s $ is forced to match the output sampled sentence $ Y_i^s $. Therefore, when the sampled sentence $ Y_i^s $ which returns higher reward than baseline is encouraged to generate, the controllability of the CIC model is also enhanced. It should be noted that due to the existence of dropout, $ p(Y_i^s|I,\beta_i) $ is probably different from $ p(Y_i^s|I,\beta_i^s) $ even when $ \beta_i $ and $ \beta_i^s $ are the same. In this situation, we tend to retain the actual sampled output distribution $ p(Y_i^s|I,\beta_i) $ instead of $ p(Y_i^s|I,\beta_i^s) $, since considering the negative impact of exposure bias. The final output distribution $ p'(Y_i^s|I,\beta_i^s) $ is updated as:
\begin{equation}\label{equ7}
p'(Y_i^s|I,\beta_i^s) = \begin{cases}
p(Y_i^s|I,\beta_i), & \beta_i = \beta_i^s \\
p(Y_i^s|I,\beta_i^s), & \beta_i \neq \beta_i^s .\\
\end{cases}
\end{equation}
After that, the CIDEr reward is calculated between $ Y_i^s $ and all ground truth captions paired with image $ I $, and finally sent into Eq. (\ref{equ6}) with $ p'(Y_i^s|I,\beta_i^s) $.

In fact, we find that samples resulting in lower rewards than baseline make the training process unstable. The reason for this phenomenon is that suppressing the output probability of sentences which are not sampled by the model itself leads to instability in the training process. Therefore, we discard the optimization of these sentences and change Eq. (\ref{equ6}) to:
\begin{equation}\label{equ8}
\nabla_\theta L_{RL}(\theta) = -\frac{1}{k}\sum_{i=1}^{k} [r(Y_i^s) - b]_+ \nabla_\theta log(p'_\theta(Y_i^s|I, \beta_i^s)),
\end{equation}
\noindent where $ [x]_+ = max(x, 0) $. Algorithm \ref{alg1} summarizes the entire process.

\begin{algorithm}[tb]
\caption{Training Procedure of SAT Method}
\label{alg1}
\textbf{Input}: given an image $ I $, paired ground truth captions $ G = \{G_1, G_2, \cdots, G_k\} $\\
\textbf{Output}: controllable captioner $ C $
\begin{algorithmic}[1] 
\FOR{epoch in $ [M, N) $}
\FOR{each ground truth caption $ G_i $ }
\STATE Calculate the control signal $ \beta_i $ based on $ G_i $.
\STATE Use a single Monte-Carlo sample to generate a caption $ Y_i^s = C(I, \beta_i) $ and its corresponding distribution $ p(Y_i^s|I,\beta_i) $.
\STATE Calculate the control signal $ \beta_i^s $ based on $ Y_i^s $.
\STATE Generate the output distribution $ p(Y_i^s|I,\beta_i^s) $.
\STATE Change the output distribution $ p(Y_i^s|I,\beta_i^s) $ to $ p'(Y_i^s|I,\beta_i^s) $ as in \cref{equ7}.
\STATE Calculate reward $ r(Y_i^s) $ based on $ Y_i^s $ and $ G $.
\ENDFOR
\STATE Optimize $ C $ with Eq. (\ref{equ8}).
\ENDFOR
\end{algorithmic}
\end{algorithm}

\subsection{Finetuning}
\label{sec:formatting42}
After the CIC model converges with self-annotated training, we propose a finetuning method to further improve the controllability of the model. Our work is inspired by the extra reward proposed in \cite{Cornia2020Show}. It is designed to evaluate the alignment with respect to the input control signal, thus enhancing the controllability of the model. However, the extra reward is so tailored for the specific task that it is difficult to imitate the construction of it on other tasks.

In this paper, we propose a general construction of reward for structure-related controllability (\Cref{fig1} (b)). Following the mathematics notation in \Cref{sec:formatting41}, $ \beta_i $ and $ \beta_i^s $ are given from steps 2 to 5 of Algorithm \ref{alg1}. We first compute their embedding vectors $ e_i = \{e_{len}, e_{ten}\} $ and $ e_i^s = \{e_{len}^s, e_{ten}^s\} $ as in \cref{equ1}. Then, the extra reward, which evaluates the alignment between the input control signal $ \beta_i $ and the attribute $ \beta_i^s $ of the output sentence, is formulated as the form of Euclidean distance:
\begin{equation}\label{equ9}
r_{align} = -\frac{\Vert e_{len} - e_{len}^s \Vert_2 + \Vert e_{ten} - e_{ten}^s \Vert_2}{2 \sqrt{d}},
\end{equation}
where d is the embedding dimension. The final reward $ r(Y_i^s) $ is a weighted sum of CIDEr score and the alignment score:
\begin{equation}\label{equ10}
r(Y_i^s) = r_{cider} + \lambda  r_{align},
\end{equation}
where $ \lambda $ is a trade-off parameter to balance the contributions between accuracy and controllability. In the finetuning stage, the CIC model is optimized by conventional reinforcement training with \cref{equ6} instead of SAT method. The generalizability of our finetuning method comes from that as long as control signals can be encoded into vectors, our method is able to evaluate the alignment reward. Under the extra supervision of the alignment reward, the controllability of CIC models is improved.

\section{Experiments}
\subsection{Datasets and Evaluation Metrics}
We use the MSCOCO 2014 captions dataset \cite{Lin2014Microsoft} to evaluate our proposed methods. MSCOCO dataset includes 164,062 images labeled with 5 captions each. Following the Karpathy data split \cite{Karpathy2016Deep} which has been widely used in prior work, we choose 113,287 for training, 5,000 images for validation and 5,000 images for test. We measure the caption quality by using five evaluation metrics, including BLEU \cite{papineni-etal-2002-bleu}, ROUGE-L \cite{lin-2004-rouge}, METEOR \cite{Denkowski2014Meteor}, CIDEr \cite{Vedantam2015CIDEr} and SPICE \cite{2016SPICE}. Following the approach in \cite{2021Human}, we adopt a new metric Control Precision (CP) to measure the alignment with the input control signal. 

\subsection{Implementation Details}
We choose AoANet \cite{Huang2019Attention} as our baseline model and follow \cite{Huang2019Attention} to set its hyper-parameters. Specifically, we extract image features by employing Faster-RCNN \cite{Ren2015Faster} pretrained on Visual Genome \cite{Krishna2017Visual}, thus obtaining a 2048-dimensional feature vector for each region. The input word embedding size and the hidden state size are all set to 1024. We adopt Adam optimizer to minimize the cross-entropy loss for 30 epochs, and then use self-annotated training with a fixed learning rate of $ 5\times10^{-6} $ for another 20 epochs. After that, AoANet is further trained with the finetuning method for 5 epochs. The batch size is set to 10 and the beam size is set to 2. The trade-off coefficient $ \lambda $ in \cref{equ10} is set to 1. 

\begin{table*}[htbp]
	\begin{center}
		\begin{tabular}{p{25 pt}<{\centering}m{15 pt}<{\centering}m{30 pt}<{\centering}|cccccc|cccccc|c}
		    \hline
		    \multicolumn{3}{c|}{Training strategy}&\multicolumn{6}{c|}{1 caption to 1 ground truth}&\multicolumn{6}{c|}{1 caption to 5 ground truth}& con. \\
			\hline
			XE/RL& SAT& Finetune & B-1 & B-4 & M & R & C & S & B-1 & B-4 & M & R & C & S & CP  \\ \hline
			\hspace{5pt}XE & \hspace{2pt}$ \times $ & $ \times $ & 45.3 & \textbf{15.9} & 19.7 & 41.8 & 147.6 & 28.2 & 74.5 & 33.8 & 27.9 & 56.1 & 111.3 & 21.3 & 97.6 \\ 
			\hspace{5pt}RL & \hspace{2pt}$ \times $ & $ \times $ & 39.6 & 14.2 & 18.6 & 42.0 & 140.7 & 28.0 & 74.0 & \textbf{36.1} & 27.0 & 57.0 & 116.4 & 20.6& 28.7  \\ 
			\hspace{5pt}RL & \hspace{2pt}$ \times $ & $ \checkmark $ & 45.8 & 15.2 & 19.8 & 42.1 & 148.0 & 28.4 & 76.8 & 34.7 & 28.2 & 56.9 & 120.3 & 21.8& \textbf{98.3} \\
			\hspace{5pt}RL & \hspace{2pt}$ \checkmark $ & $ \times $ & \textbf{45.9} & 15.6 & \textbf{19.9} & \textbf{42.2} & \textbf{150.4} & 28.6 & 77.0 & {35.0} & \textbf{28.3} & 57.0 & 120.5 & 21.8 & 97.7 \\ 
			\hspace{5pt}RL & \hspace{2pt}$ \checkmark $ & $ \checkmark $ & \textbf{45.9} & 15.4 & \textbf{19.9} & \textbf{42.2} & 149.7 & \textbf{28.7} & \textbf{77.4} & {35.0} & \textbf{28.3} & \textbf{57.1} & \textbf{121.9} & \textbf{22.0} & \textbf{98.3}\\\hline
		\end{tabular}
	\end{center}
	\caption{Ablation study on AoANet baseline model in the task of controlling sentence length and tense simultaneously, where B-1, B-4, M, R, C, S, CP and con. represent BLEU1, BLEU4, METEOR, ROUGE-L, CIDEr-D, SPICE, Control Precision and controllability respectively. All values are reported as percentage ($ \% $).}
\label{Table3}\end{table*}

\begin{table*}[htbp]
	\begin{center}
	\setlength{\tabcolsep}{5.4pt}{
		\begin{tabular}{cc|c|cccccc|cccccc|c}
		    \hline
			\multicolumn{2}{c|}{Task} & Training & \multicolumn{6}{c|}{1 caption to 1 ground truth}&\multicolumn{6}{c|}{1 caption to 5 ground truth}& con. \\
			\hline
    		Length & Tense & Method & B-1 & B-4 & M & R & C & S & B-1 & B-4 & M & R & C & S & CP \\ \hline
			\multirow{2}{*}{$ \checkmark $} & \multirow{2}{*}{$ \times $} & XE & 42.3 & 12.6 & 18.5 & 38.8 & 122.1 & 26.3 & 74.8 & 34.2 & 28.2 & 56.5 & 113.1 & 21.5 & 99.9\\
			& & SAT & 43.5 & 12.8 & 19.0 & 39.8 & 129.1 & 26.9 & 77.7 & 36.1 & 28.7 & 57.4 & 124.2 & 22.1 & 99.9 \\ \hline
			\multirow{2}{*}{$ \times $} & \multirow{2}{*}{$ \checkmark $}& XE & 42.7 & 14.8 & 19.1 & 41.2 & 139.1 & 28.4 & 77.7 & 36.5 & 27.9 & 57.0 & 116.1 & 21.2 & 98.2\\
			& & SAT & 44.4 & 15.2 & 19.7 & 42.3 & 145.1 & 29.1 & 80.1 & 38.0 & 28.8 & 58.5 & 125.7 & 22.6 & 98.8\\ \hline
			\multirow{2}{*}{$ \checkmark $} & \multirow{2}{*}{$ \checkmark $} & XE & 45.3 & 15.9 & 19.7 & 41.8 & 147.6 & 28.2 & 74.5 & 33.8 & 27.9 & 56.1 & 111.3 & 21.3 & 97.6\\
			& & SAT & {45.9} & 15.4 & {19.9} & {42.2} & 149.7 & {28.7} & {77.4} & {35.0} & {28.3} & {57.1} & {121.9} & {22.0} & {98.3}\\\hline
		\end{tabular}}
	\end{center}
	\caption{Evaluation of AoANet w/ or w/o our SAT method in different tasks.}
\label{Table4}\end{table*}

\begin{figure}[htbp]
\centering
\includegraphics[width=1\columnwidth]{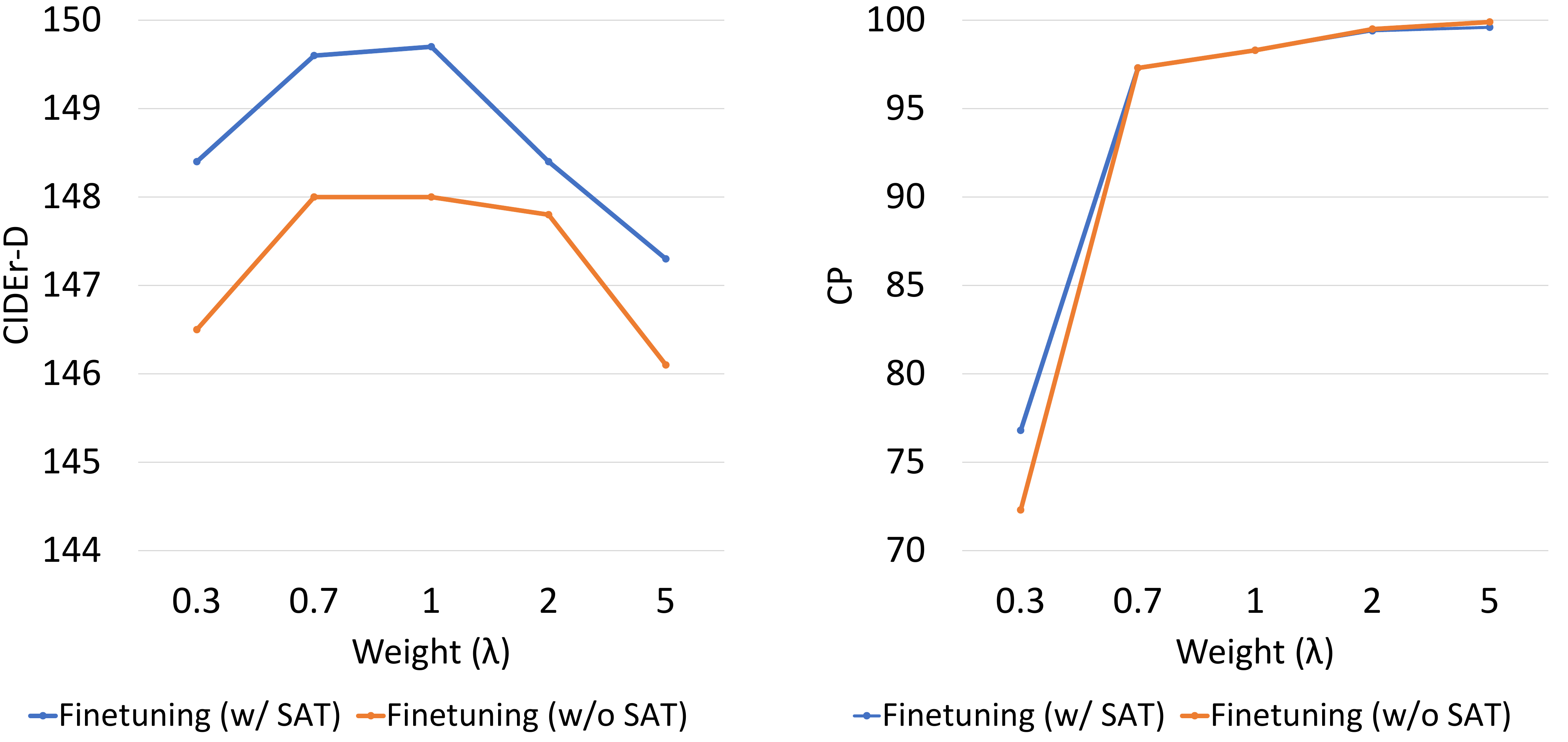} 
\caption{Performances evaluation of our finetuning method with different trade-off coefficients $ \lambda $. The left one shows the CIDEr-D score while the right one shows the control precision.}
\label{fig2}
\end{figure}

\subsection{Quantitative Analysis}
By reviewing the previous works for CIC tasks, we find that content-related works \cite{Cornia2020Show,Chen2020Say,2021Human} prefer to use `` 1 caption to 1 ground truth '' to test the results of CIC models. `` 1 caption to 1 ground truth '' means the accuracy-based metrics are calculated between the generated sentence and the single ground truth caption which provides the control signal. By contrast, structure-related works \cite{Deshpande2019Fast,2020Length} prefer to adopt `` 1 caption to 5 ground truth '', which means the accuracy-based metrics are calculated between the generated sentence and all ground truth captions aligned with the image. On the surface, `` 1 caption to 1 ground truth '' pays more attention to controllability while `` 1 caption to 5 ground truth '' pays more attention to accuracy. In order to fully demonstrate the performance of our method, we show the results in both situations. Note that in the following sections except \Cref{sec:formatting53}, the symbol `` SAT '' represents the whole process of SAT + Finetuning.

\begin{table*}[tb]
	\begin{center}
		\begin{tabular}{|m{22pt}<{\centering}|c|m{130pt}|m{130pt}|m{130pt}|}
			\hline
			\multicolumn{2}{|c|}{image} &
			\vspace{1mm}
			\includegraphics[width=129pt, height=90pt]{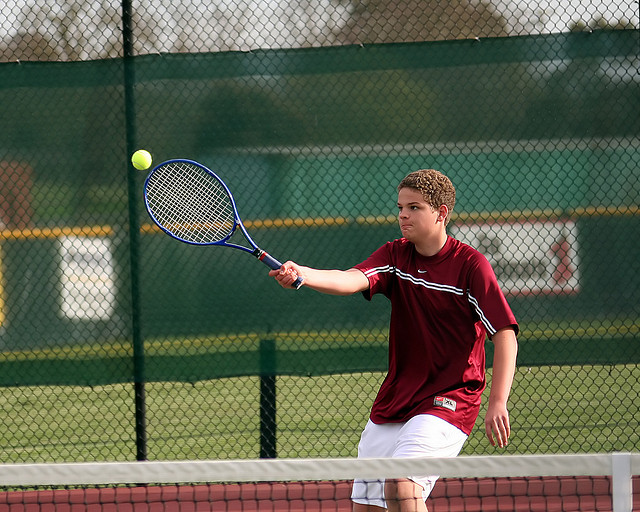} & 
			\vspace{1mm}
			\includegraphics[width=129pt, height=90pt]{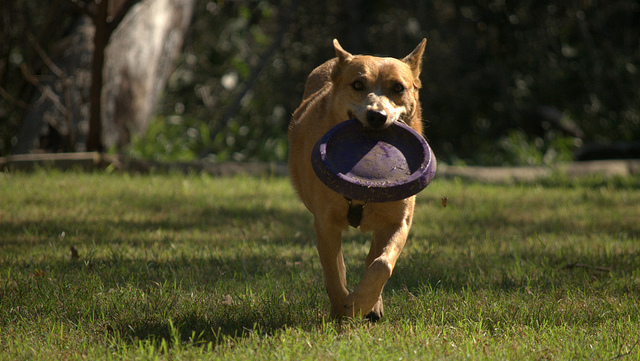} & 
			\vspace{1mm}
			\includegraphics[width=129pt, height=90pt]{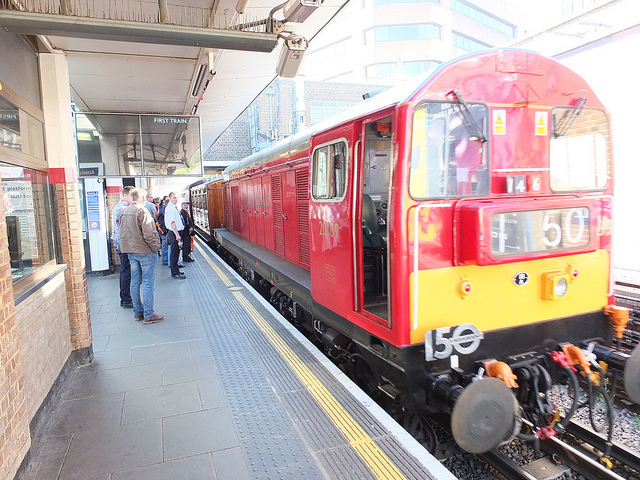} \\
			\hline
		
			& $ \leq $8 & a man playing tennis on a tennis court & a dog holding a frisbee in its mouth & a red train at a train station \\
			\cdashline{2-5}[2pt/2pt]
			& 9 & a man swinging a tennis racket at a ball & a dog is holding a frisbee in its mouth & a red and yellow train at a train station \\
			\cdashline{2-5}[2pt/2pt]
			length & 10 & a man swinging a tennis racket at a tennis ball & a dog is holding a blue frisbee in its mouth & a red and yellow train is at a train station \\
			\cdashline{2-5}[2pt/2pt]
			& 11 & a young boy swinging a tennis racket at a tennis ball & a dog holding a frisbee in its mouth in a field & a red and yellow train is parked at a train station \\
			\cdashline{2-5}[2pt/2pt]
			& $ \geq $12 & a young boy standing on a tennis court holding a tennis racket & a brown dog with a frisbee in its mouth in a field & a red and yellow train is on the tracks at a station \\
			\hline
			
			& no v & a young boy on a court with a tennis racket & a dog with a frisbee in its mouth & a red and yellow train at a train station \\
			\cdashline{2-5}[2pt/2pt]
			tense & be + v & a young boy \underline{\textit{is playing}} tennis on a tennis court & a dog \underline{\textit{is holding}} a frisbee in its mouth & a red train \underline{\textit{is parked}} at a train station \\
			\cdashline{2-5}[2pt/2pt]
			& v-ing & a young boy \underline{\textit{swinging}} a tennis racket at a tennis ball & a dog \underline{\textit{holding}} a frisbee in its mouth & a red train \textit{\underline{pulling}} into a train station \\
			\hline
		\end{tabular}
	\end{center}
	\caption{Sample results of controllability with different control signals. In order to fully present the role of each control signal, we separately train two models with different control signals to control the sentence length and tense respectively.}
\label{Table5}\end{table*}

\begin{table}[htbp]
	\begin{center}
	\setlength{\tabcolsep}{3.6pt}{
		\begin{tabular}{ccccccc}
			\hline
			Models & B-1 & B-4 & M & R & C & S \\ \hline
			UpDown (XE) & 40.8 & 12.0 & 17.6 & 37.5 & 115.4 & 25.2  \\ 
			UpDown (SAT) & 42.5 & 12.2 & 18.2 & 39.1 & 120.9 & 25.8  \\ \hline
			AoANet (XE) & 42.3 & 12.6 & 18.5 & 38.8 & 122.1 & 26.3 \\
			AoANet (SAT) & 43.4 & 12.9 & 18.9 & 39.8 & 128.5 & 26.9  \\ \hline
			Transformer (XE) & 41.0 & 12.2 & 17.9 & 37.6 & 117.8 & 25.6  \\
			Transformer (SAT) & 43.3 & 12.6 & 18.7 & 39.6 & 128.6 & 26.5 \\\hline
		\end{tabular}}
	\end{center}
	\caption{Evaluation of different baseline models w/ or w/o our SAT method in the task of sentence length.}
\label{Table6}\end{table}

\begin{table*}[ht]
	\begin{center}
		\begin{tabular}{c|cccc|cccc|cccc}
			\hline
			Metrics & S & C & M & B-4 & S & C & M & B-4 & S & C & M & B-4 \\ \hline
			Level & \multicolumn{4}{c|}{Lv 1 (1-9)} & \multicolumn{4}{c|}{Lv 2 (10-14)} & \multicolumn{4}{c}{Lv 3 (15-19)} \\ \hline
			AoANet \cite{Huang2019Attention} & 19.6 & 107.4 & 25.9 & 33.1 & 21.7 & 117.6 & 28.6 & 35.8 & 22.7 & 79.9 & 28.7 & 26.6 \\ 
			VLP \cite{2019Unified} & 18.9 & 103.0 & 25.2 & 31.8 & 21.4 & 118.7 & 28.8 & 36.0 & 22.4 & 92.5 & 29.3 & 28.4\\ 
			LaBERT \cite{2020Length} & 19.5 & 101.6 & 25.4 & 30.0 & 21.8 & 118.2 & 28.4 & 35.3 & 22.3 & 90.5 & 28.6 & 26.8 \\ \hline
			AoANet$ ^\dag $ (XE) & 19.8 & 108.2 & 26.0 & 33.3 & 21.6 & 118.0 & 28.7 & 36.3 & 22.6 & 80.7 & 28.6 & 26.7 \\
			AoANet$ ^\dag $ (SAT) & \textbf{20.3} & \textbf{115.4} & \textbf{26.4} & \textbf{35.6} & \textbf{22.9} & \textbf{128.0} & \textbf{29.3} & \textbf{37.8} & \textbf{24.0} & \textbf{93.5} & \textbf{29.7} & \textbf{29.2}\\\hline
		\end{tabular}
	\end{center}
	\caption{Performance comparisons with the models in \cite{2020Length} in the task of sentence length. $ ^\dag $ denotes the results of our trained model. The other data are from \cite{2020Length}.}
\label{Table7}\end{table*}

\subsubsection{Ablative Analysis.} 
\label{sec:formatting53}
To examine the impact of our proposed SAT method and finetuning method, we choose AoANet \cite{Huang2019Attention} as our baseline model and conduct ablation studies in the task of controlling the sentence length and tense simultaneously. The total training epochs of all RL-based methods are set to be the same. It should be noted that the CIDEr-D score contains the term of length penalty, which plays a similar role of our alignment reward. In order to better show the ability of our methods, we remove this item when calculating the CIDEr-D reward during RL-based training. After all, the length penalty is not applicable to other structure-related tasks and cannot be relied on all the time, but both our methods are.

From the Table \ref{Table3} we can observe that: 1) Trained with the RL method which is adopted in \cite{Cornia2020Show} instead of our SAT or finetuning method, the performance of CIC models is far from satisfactory in terms of controllability. Although the reward is calculated only in relation to the ground truth caption that provides the control signal, the model fails to associate the input control signal with the true meaning of structure-related attributes, obtaining only 28.7 $ \% $ control precision. Improving the accuracy-based metrics under condition of `` 1 caption to 5 ground truth '' and falling in terms of controllability prove our point: the reward of conventional RL method focuses more on contents instead of semantic structures. 2) The introduction of SAT or finetuning method brings an improvement both in accuracy and controllability. Specifically, SAT method boosts the CIDEr-D score from 147.6 / 111.3 to 150.4 / 120.5 in two evaluation conditions respectively, and slightly improves the control precision. Based on the reinforcement training which enhances the accuracy of CIC models, our finetuning method improves the control precision from 97.6 $ \% $ to 98.3$ \% $ by introducing an extra alignment reward. 3) When the two methods are combined together, the CIC model achieves both high accuracy (from 147.6 to 149.7 and 111.3 to 121.9 CIDEr-D score in `` 1 caption to 1 ground truth '' and `` 1 caption to 5 ground truth '' respectively) and controllability (from 97.6 $ \% $ to 98.3 $ \% $ control precision). The experiments above validate the effectiveness of our SAT and finetuning method.

\subsubsection{Model Selection with $ \lambda $.} In our proposed finetuning method, we combine two rewards together with a trade-off coefficient $ \lambda $ in \cref{equ10}.  Figure \ref{fig2} shows the results of our finetuning method with different $ \lambda $. It is obvious that the control precision rises with the increase of $ \lambda $, both achieving more than 99 $ \% $ control precision when the $ \lambda $ is set to 5, which verifies the effectiveness of the alignment reward in controllability. However, focusing too much on controllability reduces the accuracy of CIC models. As in Figure \ref{fig2} (a), when $ \lambda $ is too large the CIDEr-D score drops by almost 2. Since both models achieve their best CIDEr-D performances at $ \lambda = 1 $, we eventually select $ \lambda = 1 $ for our finetuning method. By comparing the results between Finetuning (w/ SAT) and Finetuning (w/o SAT), we find that the model trained after SAT method is 1.5 higher in CIDEr-D on average than that without using SAT method, which demonstrates the necessity of our SAT method.

\subsubsection{Generalizability on Different tasks.} Table \ref{Table4} reports the generalizability of our SAT method in different tasks, including sentence length, sentence tense and their combination. As it can be observed, our SAT method significantly outperforms the XE method in all accuracy-based evaluation metrics, especially in the single task. Meanwhile, our SAT method maintains the high controllability, improving the control precision compared with the XE method. The performance above prove the effectiveness and generalizability of our SAT method.

\subsubsection{Generalizability on Different Baseline Models.} Table \ref{Table6} shows the generalizability of our SAT method in different baseline models, including UpDown \cite{Anderson2018Bottom}, AoANet \cite{Huang2019Attention} and Transformer \cite{2017Attention}. Due to the limited space, we only show the results of `` 1 caption to 1 ground truth '' and leave the data of `` 1 caption to 5 ground truth '' in the supplementary material. As reported in Table \ref{Table6}, all baseline models with our SAT method outperform that with XE training in terms of all evaluation metrics. Concretely, boosting the CIDEr-D score from 117.8 to 128.6 on the Transformer baseline, verifies the advantage and generalizability of our SAT method.

\subsubsection{Comparison with Previous Works} 
\begin{figure}[htbp]
\centering
\includegraphics[width=0.95\columnwidth]{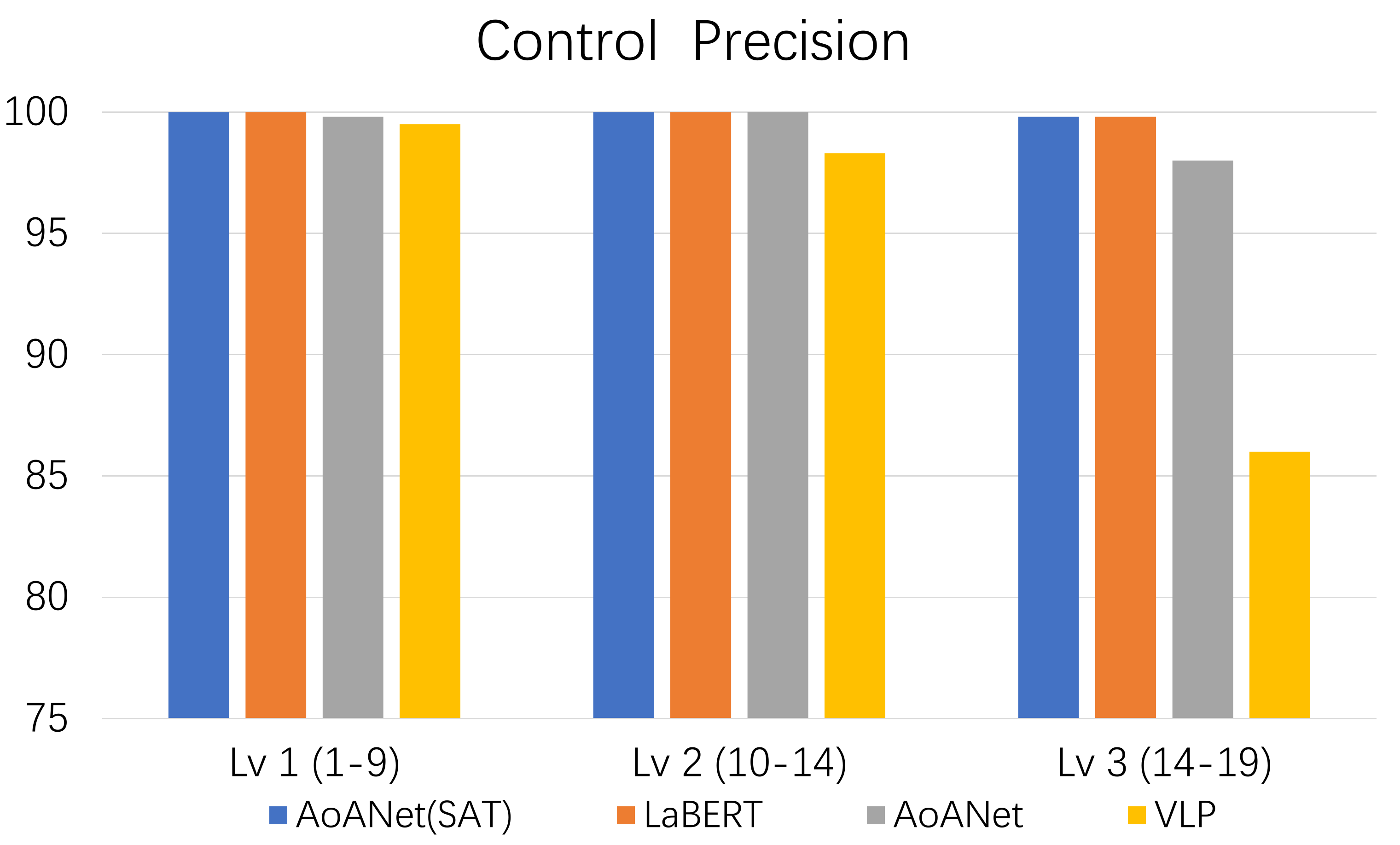} 
\caption{The control precision of our SAT version and three other versions in \cite{2020Length}.}
\label{fig3}
\end{figure}

In the field of controllable image captioning, only \cite{2020Length} and \cite{Deshpande2019Fast} focus on structure-related control signals. Since the core contribution of \cite{Deshpande2019Fast} is high diversity and fast speed, we mainly compare our SAT method with the performance of models in \cite{2020Length}. In this part, we follow the settings in \cite{2020Length} and divide the sentence length into several levels: [1, 9], [10, 14] and [15, 19]. In the test stage, the input control signal is artificially fixed at a certain level. Table \ref{Table7} shows the performance comparisons between the previous works and our proposed approach in the task of sentence length. As it can be observed, AoANet trained by our SAT method achieves the best performance according to all metrics at all three levels. Boosting all evaluation metrics on the AoANet (XE) baseline, validates that our SAT method is able to greatly improve the accuracy of CIC models. \Cref{fig3} reports the control precision of the above methods. As we can see, AoANet equipped with our SAT method reaches almost the same controllability as the previously best model LaBERT \cite{2020Length}. They both achieve more than 99 $\%$ control precision at all three levels, and fully satisfy the needs of CIC tasks. The results above prove that our SAT method significantly improves the accuracy-based performance of CIC models while maintaining high controllability.

\subsection{Qualitative Analysis.} Table \ref{Table5} shows some examples controlled by different requirements of sentence attributes. Due to the limited space, we only show part of the tense level and leave the whole table in the supplementary material. When faced with different control signals, the CIC model trained with our SAT method is able to generate various captions for the same image according to the demand. Take the performance in the length task as an example, with the increase of the control level, the length of the generated sentence also increases word by word. With our self-annotated training, the generated captions achieve both high accuracy and controllability, which illustrates the effectiveness of our methods.

\section{Conclusion}
In this paper, we focus on the reinforcement training method for controllable image captioning. For structure-related control signals, we propose a novel reinforcement training method called SAT, which adopts a recursive annotation mechanism to force the input control signal to match the output sentence. Moreover, we propose an extra alignment reward to finetune the CIC model trained after SAT method. Extensive experiments validate the effectiveness of our methods in terms of accuracy and controllability.

{
\bibliography{aaai22}
}

\end{document}


\maketitle

\begin{abstract}
The Controllable Image Captioning (CIC) task aims to generate captions conditioned on designated control signals. In this paper, we improve CIC from two aspects: 1) Existing reinforcement training methods are not applicable to structure-related CIC models due to the fact that the accuracy-based reward focuses mainly on contents rather than semantic structures. The lack of reinforcement training prevents the model from generating more accurate and controllable sentences. To solve the problem above, we propose a novel reinforcement training method for structure-related CIC models: Self-Annotated Training (SAT), where a recursive sampling mechanism (RSM) is designed to force the input control signal to match the actual output sentence. Extensive experiments conducted on MSCOCO show that our SAT method improves C-Transformer (XE) on CIDEr-D score from 118.6 to 130.1 in the length-control task and from 132.2 to 142.7 in the tense-control task, while maintaining more than 99$\%$ matching accuracy with the control signal. 2) We introduce a new control signal: sentence quality. Equipped with it, CIC models are able to generate captions of different quality levels as needed. Experiments show that without additional information of ground truth captions, models controlled by the highest level of sentence quality perform much better in accuracy than baseline models.
\end{abstract}

\section{Introduction}

control signals aligned with ground truth captions are used as inputs during reinforcement training. Under the guidance of them, the content of the generated sentence is trained to approach the ground truth caption, so that CIC models are able to learn content-related controllability. However, this method can only be applied to content-related control signals because the accuracy-based reward in reinforcement learning is designed to measure the content similarity between two sentences. For structure-related control signals, the accuracy-based reward makes the generated sentence approach the ground truth sentence in contents rather than semantic structures. Therefore, CIC models fail to learn structure-related controllability with ground truth annotations in reinforcement training.

By reviewing the cross-entropy training method, we find that the main reason why models can learn controllability well is the consistency of inputs and outputs, that is, the input control signal can match the output sentence of models (in cross-entropy training output sentences of models are ground truth captions). In this case, CIC models will naturally associate the input control signal with the output sentence, thus learning the meaning of the control signal. Therefore, in reinforcement training, control signals should be aligned with actual output sentences which are generated by sample methods rather than ground truth captions (Fig.). However, output sentences generated by sample methods require control signals as the input. The control signals, in turn, are calculated based on output sentences. They are prerequisite to each other, which brings great obstacles to the actual operation. Therefore, how to design an algorithm to solve the above problem is crucial.

In this paper, we propose a novel reinforcement training method, \emph{Self-Annotated Training} (SAT), for structure-related control signals and prove that SAT can improve the performance of CIC models dramatically in terms of accuracy and controllability. SAT is based on self-critical sequence training (SCST) method. The main difference between SAT and the method in is the source of control signals. In reinforcement training, control signals in  come from ground truth captions, while ours are from generated sentences of CIC models. In order to solve the problem that the input and output are prerequisite to each other, we design a recursive sampling mechanism (RSM) which forces the input control signal to match the output sentence. The results of RSM are then sent into the REINFORCE algorithm for training. Experiments on MSCOCO dataset  demonstrate that SAT can effectively improve the accuracy and controllability of captioning models controlled by structure-related signals.

As another contribution of this paper, we introduce a new control signal: sentence quality, which is calculated according to accuracy-based evaluation metrics. Through cross-entropy training or our self-annotated training, captioning models are able to generate sentences of different quality levels as needed. Extensive experiments on mainstream captioning models show that, without additional information of ground truth captions, models controlled by the highest quality level surpass baseline models a lot on accuracy-based evaluation metrics.

\section{Supplementary Material}
\subsection{Sentence Attribute} Table \ref{Table1} and \ref{Table2} show the specific settings of control levels for sentence length and tense respectively. We set a level for each length from 7 to 14. Since sentences less than 7 words or greater than 14 words are relatively rare in the dataset, we set a level for each of these two parts. In total, ten levels are set for sentence length. In terms of sentence tense, we set five levels to represent no verb, be + v, v-ing, v and v-ed respectively.

\subsection{Sentence Quality} Table \ref{Table3} illustrates the classification of control levels for sentence quality. The number of quality levels is set to 3 and 5 for UpDown-based models and Transformer-based models respectively due to their final performance. It is worth noting that the distribution of generated sentences from models is different from that of ground truth captions. Therefore, the threshold between levels for generated sentences and ground truth captions are different.

\begin{table}[htbp]
	\begin{center}
		\begin{tabular}{|c|c|}
			\hline
			level & length \\
			\hline
			0 & $ \textless $ 7	\\ \hline
            1 & $ = $ 7	\\ \hline
            2 & $ = $ 8 \\	 \hline
            3 & $ = $ 9 \\ \hline
            4 & $ = $ 10	\\ \hline
			5 & $ = $ 11	\\ \hline
			6 & $ = $ 12	\\ \hline
			7 & $ = $ 13\\ \hline
			8 & $ = $ 14	\\ \hline
			9 & $ \textgreater $ 14	\\ \hline
		\end{tabular}
	\end{center}
	\caption{Specific settings of control levels for sentence length.}
\label{Table1}\end{table}

\begin{table}[htbp]
	\begin{center}
		\begin{tabular}{|c|c|}
			\hline
			level & tense \\
			\hline
			0 & no v	\\ \hline
            1 & be + v	\\ \hline
            2 & v-ing \\	 \hline
            3 & v \\ \hline
            4 & v-ed	\\ 
			\hline
		\end{tabular}
	\end{center}
	\caption{Specific settings of control levels for sentence tense.}
\label{Table2}\end{table}

\begin{table}[htbp]
	\begin{center}
		\begin{tabular}{|c|c|c|}
			\hline
			level & UpDown & Ground Truth\\
			\hline
			0 & $ x \textless 0.5 $ & $ x \textless 0.375 $	\\ \hline
            1 & $ 0.5 \leq x \textless 0.9 $ & $ 0.375 \leq x \textless 0.625 $	\\ \hline
            2 & $ 0.9 \leq x \textless 1.3 $ & $ 0.625 \leq x \textless 0.875 $ \\	 \hline
            3 & $ 1.3 \leq x \textless 1.7 $ & $ 0.875 \leq x \textless 1.25 $ \\ \hline
            4 & $ x \geq 1.7 $ & $ x \geq 1.25 $	\\ \hline
            \hline
            level & Transformer & Ground Truth\\
			\hline
			0 & $ x \textless 0.7 $ & $ x \textless 0.375 $	\\ \hline
            1 & $ 0.7 \leq x \textless 1.3 $ & $ 0.375 \leq x \textless 0.625 $	\\ \hline
            2 & $ x \geq 1.3 $ & $ x \geq 0.625 $ \\	 \hline
		\end{tabular}
	\end{center}
	\caption{Specific settings of control levels for sentence quality. The second column is for reinforcement training, and the third column is for cross-entropy training.}
\label{Table3}\end{table}